# OPTIMIZED METHOD FOR IRANIAN ROAD SIGNS DETECTION AND RECOGNITION SYSTEM

Reza Azad[1], Babak Azad[2], Iman Tavakoli Kazerooni[3]

[1]Ardabil Branch Islamic Azad University, Ardabil, IRAN
Email: rezazad68@gmail.com

[2]Institute of Computer science, Shahid Bahonar University, Shiraz, IRAN
Email: babak.babi72@gmail.com

[3]Department of Computer Engineering Hamedan Branch, Islamic Azad University, Science and Research Campus, Hamedan, IRAN
Email: iman_tavakoli2008@yahoo.com

*Abstract:* Road sign recognition is one of the core technologies in Intelligent Transport Systems. In the current study, a robust and real-time method is presented to identify and detect the roads speed signs in road image in different situations. In our proposed method, first, the connected components are created in the main image using the edge detection and mathematical morphology and the location of the road signs extracted by the geometric and color data; then the letters are segmented and recognized by Multiclass Support Vector Machine (SVMs) classifiers. Regarding that the geometric and color features ate properly used in detection the location of the road signs, so it is not sensitive to the distance and noise and has higher speed and efficiency. In the result part, the proposed approach is applied on Iranian road speed sign database and the detection and recognition accuracy rate achieved 98.66% and 100% respectively.

*Keywords:* Road sign detection and recognition, edge detection, mathematical morphology, SVM classifier.

## I. INTRODUCTION

Road sign recognition is one of the core technologies in Intelligent Transport Systems (ITS). This is due to the importance of the road signs and traffic signals in daily life. They define a visual language that can be interpreted by the drivers. They represent the current traffic situation on the road, show the danger and difficulties around the drivers, give warnings to them, and help them with their navigation by providing useful information that makes the driving safe and convenient [1].

The field of road sign recognition is not very old; the first paper appeared in Japan in 1984. The aim was to try various computer vision methods for the detection of objects in outdoor scenes. Since that time many research groups and companies are interested and conducted research in the field, and enormous amount of work has been done. Different techniques have been used, and big improvements have been achieved during the last decade.

Road sign recognition systems usually have developed into two specific phases [2-9]. The first is normally related to the detection of traffic signs in a video sequence or image using image processing. The second one is related to recognition of these detected signs. The detection algorithms normally based on shape or color segmentation. The prevalent approach in detecting traffic signs based on color is very obvious one finds the areas of the image which contain the color of interest, using simple thresholding or more advanced image segmentation methods. The resulting areas are then either immediately designated as traffic signs, or passed on to subsequent stages as traffic sign location hypotheses (i.e. regions of interest). The main weakness of such an approach lies in the fact that color tends to be unreliable - depending on the time of day, weather conditions, shadows etc. the illumination of the scene can vary considerably. RGB color space is considered to be very sensitive to illumination, so many researchers choose to carry out the color-based segmentation in other color spaces, such as HSI or L*a*b. Some methods that implemented on this approach are [10-20].

Several approaches for shape-based detection of traffic signs are implemented. Probably the most common approach is using some form of Hough transform. Approaches based on corner detection followed by reasoning or approaches based on simple template matching are also popular. Sample of these approaches proposed in [21-23]. Also some approaches use the prior knowledge of the problem (the expected color and shape of a traffic sign) as machine learning way for traffic sign recognition like [24].

In this paper we proposed a method based on color and shape characteristic of traffic speed sign for In section two the proposed methods is presented. In section three the practical result and implementation phase is described. Finally in section four, conclusion and future work is demonstrated.

## II. PROPOSED METHOD

General diagram of the proposed method is shown in Figure 1. In the proposed method, the input image is pre-processed at the first time and then traffic sign location is specified by edge detection and morphology operation and finally traffic sign location is extracted and their characters are recognized.

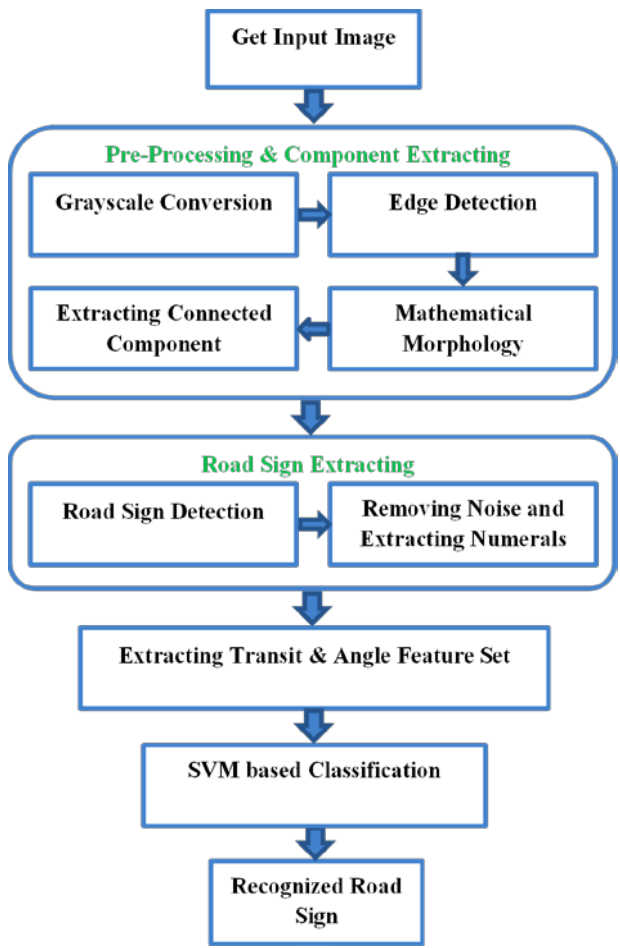

*Figure 1: General Diagram of the proposed method*

### A. Pre-processing & Component Extracting

Pre-processing is carried out on the image to improve the quality of the image so that the main processing on the image becomes easier. This step involves image converting to grey scale, edge detection, noise removing by mathematical morphology and extracting connected component.

1. Gray Scale Conversion

The road sign location in this paper are based on gray image, so the main function of the pretreatment algorithm is to convert color images to gray scale images for the latter operation. A color bitmap is composed of R, G and B 3 components. If it is a 24-bit true color image, every point is made up of three bytes which respectively represent R, G and B. Therefore, the following is color information of image. Figure 2(b) shows the entrance image in gray mode.

2. Edge Detection

Edge detection is a type of image segmentation techniques which determines the presence of an edge or line in an image and outlines them in an appropriate way [25]. The main purpose of edge detection is to simplify the image data in order to minimize the amount of data to be processed [26]. Generally, an edge is defined as the boundary pixels that connect two separate regions with changing image amplitude attributes such as different constant luminance and tristimulus values in an image [25], [27] and [28]. There are different approaches and algorithm to find out the edge in image processing that, in the meantime, canny operator due to high accuracy and low processing volume has a more favorable performance compared to other methods for our database. Figure 2(c) shows the result of edge detection on entrance image.

3. Mathematical Morphology

Mathematical morphology is the branch of image processing that argues about shape and appearance of abject in images. The erosion and dilation operators are basically operators of mathematical morphology that are used in this part to improve the edge detection image. At this step, first erosion action is applied in the edge detection image. After erosion action on image, the dilation action is done.

4. Extracting Connected Component

For extracting connected component in this step, first, the holes are filled then the area with 8 connected neighbours is labeled as connected component. Figure 2(d) shows the result.

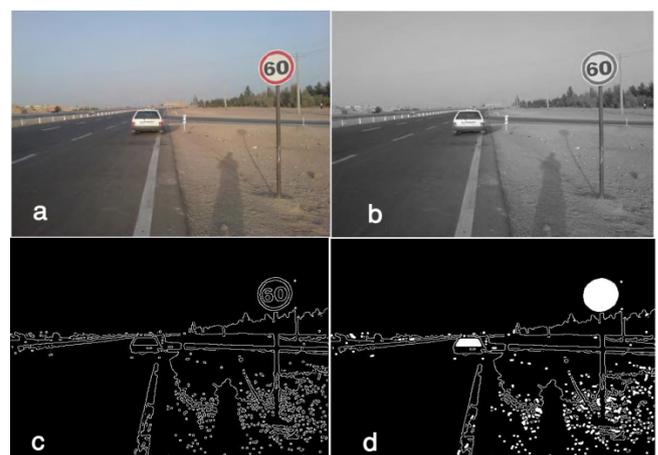

*Figure 2: a: Original image b: image in grayscale mode c: edge detected image d: extracted connected component*

*B. Road Sign Extracting*

In this stage the road sign is extracted and its characters are distinguished. This stage involves road sign detection and probability noise removing for character extracting.

1. Road Sign Detection

In this part, the color and texture features used to identify the location of the road signs. Considering that the road speed signs are circular and marked with red areas in the edge of the circle (Figure 2(a)), so these features are used in the current study to identify the location of the road speed signs. Therefore, first the degree of the circularity of every connected region is calculated using Equation (1):

$$Metric = 4\pi * Area / Perimeter^2 \qquad (1)$$

Where $\pi = 3.1415$, Area is an object area and perimeter is an object perimeter. The regions which circularity degree is between $0.9 \leq Metric \geq 1$ are chosen as candidates and the red color is examined on that region due the red edge, that region is considered as the location of the road sign and is extracted from the main image. Figure 3 shows the result.

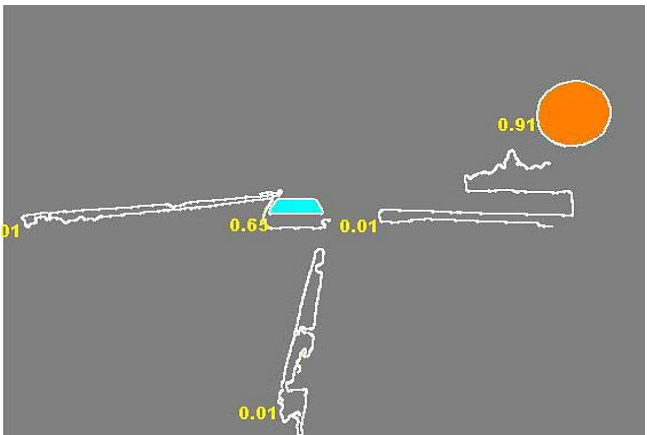

*Figure 3: a: corresponding metric of each object*

To detect the red edge, first the candidate location from the main image is brought in the colored place of HSV. then, its red pixels are detected by Equation (2). Figure 4(a) shows the extracted road sign.

$$\text{Pixel Color}(x) = \begin{cases} \text{Is Red,} & S \geq 0.45 \wedge V \geq 0.5 \wedge 0.8 \leq H \leq 0.94 \\ \text{Not Red,} & O.W \end{cases} \qquad (2)$$

2. Removing Noise and Extracting Numerals

In this paper from last step, the area that is exploited as a road sign, first probable noising and red color pixel are resolved, then road sign image is complemented till its writing of plate inside is seen such white violence. Then this area is labeled and through the available regions, the regions that are bigger are stored as exploited characters in 60*30 sizes. Figure 4(b) shows the extracted road sign after removing noise.

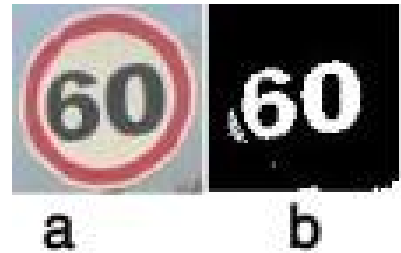

*Figure 4: a: Extracted road sign b: road sign after removing noise and red color*

*C. Feature Extracting*

In our system we computed features based on transit and angel of contour pixels of the images as follows: First we found the bounding box (minimum rectangle containing the numeral) of each input image which is a two-tone image. Then for better result and independency of features to size, font and position (invariant to scale and translation), we converted each image (located in bounding box) to a normal size of 60×30 pixels. We chose this normalized value based of various experiments and a statistical study. In Figure 5(a), a normalized image with its bounding box is shown. We extracted the contour of the normalized image Figure 5(b).

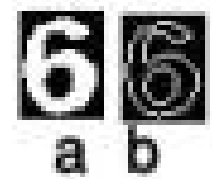

*Figure 5: a: Bounding box of a normalized image (b): Digit '6' contour form.*

We scanned the image contour horizontally by keeping a window-map of size 10×10 on the image from the top left most point to down right most point (18 non overlapped blocks). For each block the transit and angel feature were computed. To extract features, we considered 18 (10*10) uniform blocks in each image and we computed three features in each block so we got 18×2=36 features for each image.

1. Angel Features Extraction

Angle features are very important features in order to achieve higher recognition accuracy and reducing misclassification. These features are extracted from image by the equation (3) that we mentioned earlier in [29].

$$(a_b) = \frac{1}{n} + \sum_{k=1}^{n_b} \theta_k^b \qquad , b=1,2,3,.. \qquad (3)$$

In the top relation, $a_b$ is angel average for any block and $\theta_k^b$ angle of white pixel to block horizontal level. The steps that have been used to extract these features are given below [29]:

Step 1. Divide the input image into n (n=18) number of block, each of size 10×10 pixels;

Step 2. Calculate for each block of image, angel degree with use of equation (3) and set These 18 sub-features as an angel feature;

Step 3. Corresponding to the blocks whose angel does not have a foreground pixel, the feature value is taken as zero.

Using this algorithm, we will obtain 18 features corresponding to every block.

2. Transit Feature

In a binary image, whenever a pixel value changes from 0 to 1 or 1 to 0 it indicates the information about the edge. This information is very significant as it denotes the geometry of the character and helps in identifying the character [30]. In order to capture this information, we have used Run Length Count (RLC) technique. In the proposed method, for every block, we find the Run Length count in horizontal and vertical direction. A total of 18 features will be extracted for each characters and this will serve as feature vector. Following steps have been implemented for extracting these features [31].

Step 1. Divide the input image into n (n=18) number of block, each of size 10×10 pixels;

Step 2. Calculate for each block of image, Run Length count in horizontal and vertical direction and set their ratio as feature;

Step 3. Corresponding to the blocks whose does not have a foreground pixel, the feature value is taken as zero.

Using this algorithm, we will obtain 18 features corresponding to every block.

*D. Classification*

Support vector machines (SVMs) [32] are very popular and powerful in pattern learning because of supporting high dimensional data and at the same time, providing good generalization properties. Moreover, SVMs have many usages in pattern recognition and data mining applications such as text categorization [33] and [34] phoneme recognition [35], face recognition [36] and etc. At the beginning, SVM was formulated for two-class (binary) classification problems. The extension of this method to multi-class problem is neither straightforward nor unique. DAG SVM [37] is one of the methods that have been proposed to extend SVM classifier to support multi-class classification.

1. Binary support vector machine formulation

$X = \{(x_i, y_i)\}^n i = 1$ be a set of n training samples, where $x_i \in R^m$ is an m-dimensional sample in the input space, and $y_i \in \{-1, 1\}$ is the class label of sample xi. SVM finds the optimal separating hyper plane (OSH) with the minimal classification errors. The linear separation hyper plane is in the form of Equation (4).

$$f(x) = W^T x + b \qquad (4)$$

Where w and b are the weight vector and bias, respectively. The optimal hyper plane can be obtained by solving the optimization problem (7), where $\zeta_i$ is slack variable for obtaining a soft margin while variable C controls the effect of the slack variables. Separation margin increases by decreasing the value of C. In a support vector machine, the optimal hyper plane is obtained by maximizing the generalization ability of the SVM. However if the training data are not linearly separable, the obtained classifier may not have high generalization ability, even though the hyper planes are determined optimally. To enhance linear severability, the original input space is mapped into a high-dimensional do product space called the feature space. Now using the nonlinear vector function $\varphi(x) = (\varphi_1(x), \dots, \varphi_l(x))^\Psi$ that maps the m-dimensional input vector x into the l-dimensional feature space, the OSH in the feature space is given by Equation (5):

$$f(x) = W^T \varphi(x) + b \qquad (5)$$

The decision function for a test data is Equation (6):

$$D(x) = Sign(W^T \varphi(x) + b) \qquad (6)$$

The optimal hyper plane can be found by solving the following quadratic optimization problem:

$$Minimize\ \frac{1}{2}||W||^2 + C \sum_{i=1}^{n} \zeta_i$$
$$Subject\ to\ y_i(W^T \varphi(x) + b) \geq 1 - \zeta_i$$
$$\zeta_i \geq 0, i = 1, \dots, n \qquad (7)$$

2. Multiclass support vector machine

As described before, SVMs are intrinsically binary classifiers, but, the classification of faces involves more than two classes. In order to face this issue, a number of multiclass classification strategies can be adopted [38] and [39]. The most popular ones are the one-against-all (OAA) and the one-against-one (OAO) strategies. The one-against-one constructs (n(n-1))⁄2decision functions for all the combinations of class pairs. Experimental results indicate that the one-against-one is more suitable for practical use. We use OAO for extracted numeral classification.

### III. IMPLEMENTATION AND PRACTICAL RESULT

Our suggestive method have been done on Intel Core i3-2330M CPU, 2.20 GHz with 2 GB RAM under Matlab environment. Figure 6 shows the face of worked systems.

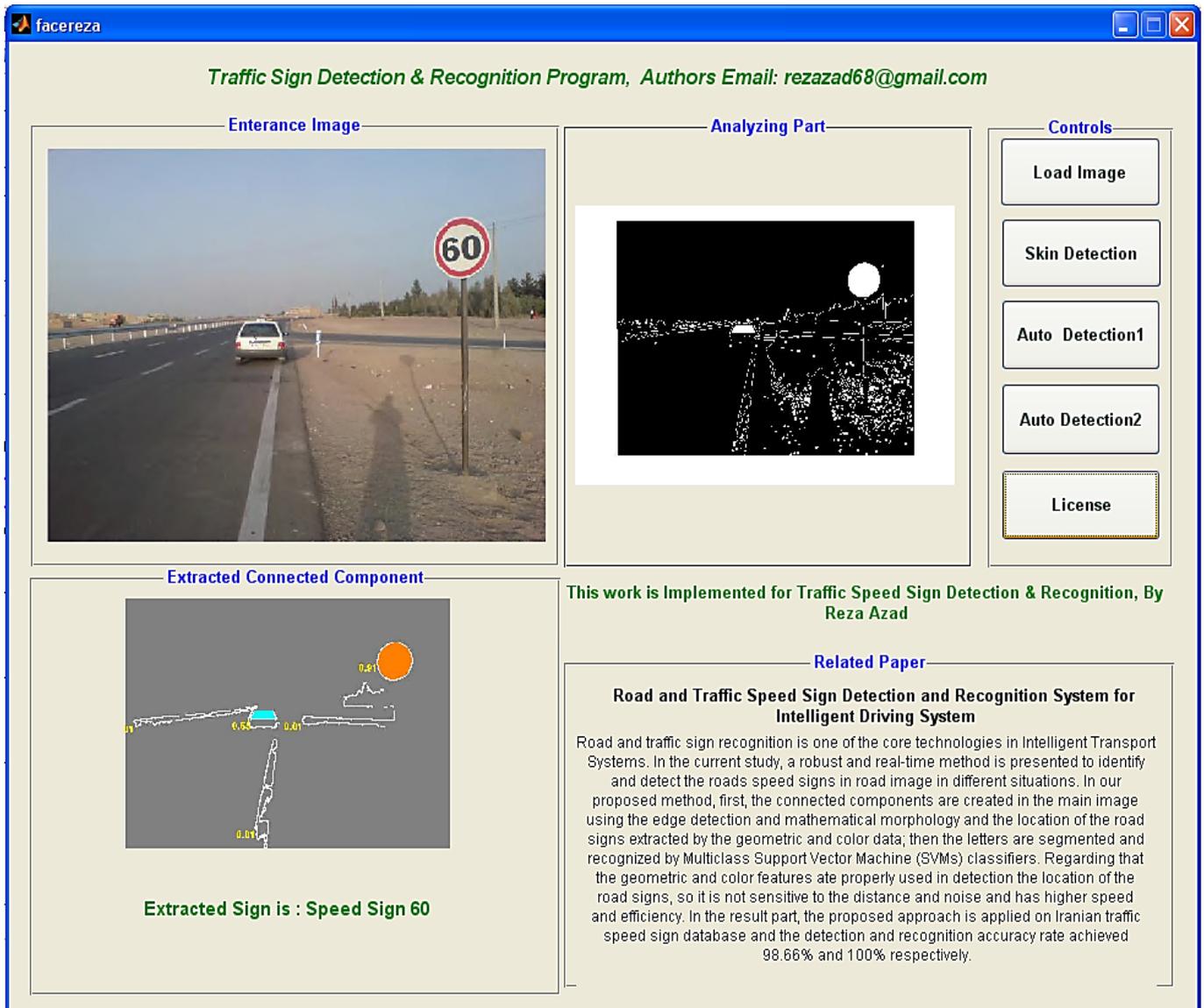

*Figure 6: road speed sign detection and recognition system*

In this study, for experimental analysis, we considered an Iranian road speed sign database for detection and recognition. The Iranian road speed sign database is an Iranian database that contains a set of 375 colorful traffic images with 5 classes for each limit of speeds [40]. These images have different background, distance observes and angel of view, also some of them contain noises and categorized in 5 class with limit speed from 20 to 100 by English writing digits. The results are as presented in table 1 and 2.

*Table 1: Comparison of different algorithms on traffic sign detection*

| Number of Images | Iranian Road Speed Sign Data Set | | |
|---|---|---|---|
| | Method | Technique | Efficiency |
| 375 | [40] | Normalized cross correlation | 93% |
| | Our Method | Combination of Shape and Color feature | 98.66% |

*Table 2: Comparison of different algorithms on Road sign Recognition*

| Number of Images | Iranian Road Speed Sign Data Sset | | |
|---|---|---|---|
| | Method | Technique | Efficiency |
| 375 | [40] | Normalized cross correlation | 93% |
| | Our Method | SVM Classifier | 100% |

Our method compared with [40] has best result because it's independent on noise and distance. Figure 7 shows the accuracy rate in different distance betwixt our method and [40]. Blue line shows our method accuracy and red line shows the [40] method accuracy in each distance.

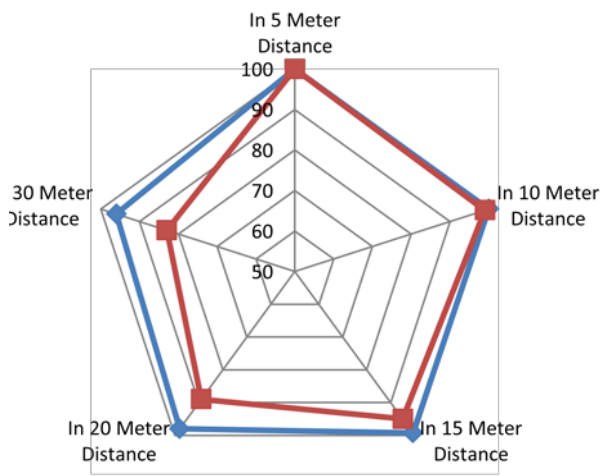

*Figure 7: Accuracy rate in different distances*

High detection rate shows the quality of proposed approach to use in every applications, which are needed a road speed sign detection and recognition stage. Low complexity in computation and time are some of other advantages of the proposed approach.

## IV. CONCLUSION AND FUTURE WORK

In this paper we propose an approach based on shape and color characteristic for road speed sign detection and recognition in still images. In our mentioned method, first, the connected components are created in the main image using the edge detection and mathematical morphology and the location of the road signs extracted by the geometric and color data; then the letters are segmented and recognized by Multiclass Support Vector Machine (SVMs) classifiers. We tested our method on Iranian road speed sign database and the accuracy rate reached 98.66 and 100% for detection and recognition stage respectively. Regarding that the geometric and color features ate properly used in detection the location of the road signs, so it is not sensitive to the distance and noise and has higher speed and efficiency.

This work is implemented for still images, for future work we have planned to extend it for road sign detection and recognition in video stream. Also occasionally accuracy decreased when the background was so complex, for solving this problem in the future work we will express the method based on edge analysis stage for reducing complexity and solving this problem.

## V. REFERENCES

[1] C. Fang, S. Chen, and C. Fuh, "Road-sign detection and tracking," IEEE Trans. on Vehicular Technology, vol. 52, pp. 1329-1341, 2003.

[2] R. Vicen-Bueno, R. Gil-Pita, M.P. Jarabo-Amores and F. L´opez-Ferreras, "Complexity Reduction in Neural Networks Applied to Traffic Sign Recognition", Proceedings of the 13th European Signal Processing Conference, Antalya, Turkey, September 4-8, 2005.

[3] R. Vicen-Bueno, R. Gil-Pita, M. Rosa-Zurera, M. Utrilla-Manso, and F. Lopez-Ferreras, "Multilayer Perceptrons Applied to Traffic Sign Recognition Tasks", LNCS 3512, IWANN 2005, J. Cabestany, A. Prieto, and D.F. Sandoval (Eds.), Springer-Verlag Berlin Heidelberg 2005, pp. 865-872, doi: 10.1007/11494669_106

[4] H. X. Liu, and B. Ran, "Vision-Based Stop Sign Detection and Recognition System for Intelligent Vehicle", Transportation Research Board (TRB) Annual Meeting 2001, Washington, D.C., USA, January 7-11, 2001, doi: 10.3141/1748-20

[5] H. Fleyeh, and M. Dougherty, "Road And Traffic Sign Detection And Recognition", Proceedings of the 16th Mini - EURO Conference and 10th Meeting of EWGT, pp. 644-653.

[6] D. S. Kang, N. C. Griswold, and N. Kehtarnavaz, "An Invariant Traffic Sign Recognition System Based on Sequential Color Processing and Geometrical Transformation", Proceedings of the IEEE Southwest Symposium on Image Analysis and Interpretation Volume , Issue , 21-24 Apr 1994, pp. 88 – 93, doi: 10.1109/IAI.1994.336679

[7] M. Rincon, S. Lafuente-Arroyo, and S. Maldonado-Bascon, "Knowledge Modeling for the Traffic Sign Recognition Task", Springer Berlin / Heidelberg Volume 3561/2005, pp. 508-517, doi: 10.1007/11499220_52

[8] C. Y. Fang, C. S. Fuh, P. S. Yen, S. Cherng, and S. W. Chen,"An Automatic Road Sign Recognition System based on a Computational Model of Human Recognition Processing", Computer Vision and Image Understanding, Vol. 96 , Issue 2 (November 2004), pp. 237 – 268, doi: 10.1016/j.cviu.2004.02.007

[9] C. Bahlmann, Y. Zhu, V. Ramesh, M. Pellkofer, T. Koehler, "A System for Traffic Sign Detection, Tracking, and Recognition Using Color, Shape, and Motion Information", Proceedings of the 2005 IEEE

Intelligent Vehicles Symposium, Las Vegas, USA., June 6 - 8, 2005, doi: 10.1109/IVS.2005.1505111

[10] M. Benallal and J. Meunier, "Real-time color segmentation of road signs," Electrical and Computer Engineering, 2003. IEEE CCECE 2003. Canadian Conference on, vol. 3, pp. 1823–1826 vol.3, May 2003, doi: 10.1109/CCECE.2003.1226265

[11] L. Estevez and N. Kehtarnavaz, "A real-time histographic approach to road sign recognition," Image Analysis and Interpretation, 1996., Proceedings of the IEEE Southwest Symposium on, pp. 95–100, Apr 1996, doi: 10.1109/IAI.1996.493734

[12] S. Varun, S. Singh, R. S. Kunte, R. D. S. Samuel, and B. Philip, "A road traffic signal recognition system based on template matching employing tree classifier," in ICCIMA '07: Proceedings of the International Conference on Computational Intelligence and Multimedia Applications (ICCIMA 2007). Washington, DC, USA: IEEE Computer Society, 2007, pp. 360–365, doi: 10.1109/ICCIMA.2007.190

[13] A. Broggi, P. Cerri, P. Medici, P. Porta, and G. Ghisio, "Real time road signs recognition," Intelligent Vehicles Symposium, 2007 IEEE, pp. 981–986, June 2007, doi: 10.1109/IVS.2007.4290244

[14] A. Ruta, Y. Li, and X. Liu, "Detection, tracking and recognition of traffic signs from video input," Oct. 2008, pp. 55–60, doi: 10.1109/ITSC.2008.4732535

[15] W.-J. Kuo and C.-C. Lin, "Two-stage road sign detection and recognition," Multimedia and Expo, 2007 IEEE International Conference on, pp. 1427–1430, July 2007, doi: 10.1109/ICME.2007.4284928

[16] G. Piccioli, E. D. Micheli, P. Parodi, and M. Campani, "Robust method for road sign detection and recognition," Image and Vision Computing, vol. 14, no. 3, pp. 209–223, 1996, doi: 10.1016/0262-8856(95)01057-2

[17] P. Paclík, J. Novovičová, P. Pudil, and P. Somol, "Road sign classification using laplace kernel classifier," Pattern Recogn. Lett., vol. 21, no. 13-14, pp. 1165–1173, 2000.

[18] C.-Y. Fang, S.-W. Chen, and C.-S. Fuh, "Road-sign detection and tracking," vol. 52, no. 5, pp. 1329–1341, Sep. 2003.

[19] A. D. L. Escalera, J. M. A. Armingol, and M. Mata, "Traffic sign recognition and analysis for intelligent vehicles," Image and Vision Computing, vol. 21, pp. 247–258, 2003.

[20] X. Gao, L. Podladchikova, D. Shaposhnikov, K. Hong, and N. Shevtsova, "Recognition of traffic signs based on their colour and shape features extracted using human vision models," Journal of Visual Communication and Image Representation, vol. 17, no. 4, pp. 675–685, 2006, doi: 10.1016/j.jvcir.2005.10.003

[21] G. Loy, "Fast shape-based road sign detection for a driver assistance system," in In IEEE/RSJ International Conference on Intelligent Robots and Systems (IROS, 2004, pp. 70–75, doi: 10.1109/IROS.2004.1389331

[22] C. Paulo and P. Correia, "Automatic detection and classification of traffic signs," in Image Analysis for Multimedia Interactive Services, 2007. WIAMIS '07. Eighth International Workshop on, June 2007, doi: 10.1109/WIAMIS.2007.24

[23] D. Gavrila, "Traffic sign recognition revisited," in DAGM-Symposium, 1999, pp. 86–93, doi: 10.1007/978-3-642-60243-6_10

[24] K. Brkić, A. Pinz, and S. Šegvić, "Traffic sign detection as a component of an automated traffic infrastructure inventory system," Stainz, Austria, May 2009.

[25] W. Frei and C. Chen, "Fast Boundary Detection: A Generalization and New Algorithm," IEEE Trans. Computers, vol. C-26, no. 10, pp. 988-998, Oct. 1977, doi: 10.1109/TC.1977.1674733

[26] J. Canny, "A computational approach to edge detection," IEEE Trans. Pattern Analysis and Machine Intelligence, Vol. 8, No. 6, pp. 679-698, Nov. 1986, doi: 10.1109/TPAMI.1986.4767851

[27] R. C. Gonzalez and R. E. Woods, Digital Image Processing. Upper Saddle River, NJ: Prentice-Hall, 2011, pp. 572-585, doi: 10.1088/1742-6596/332/1/012031

[28] W. K. Pratt, Digital Image Processing. New York, NY: Wiley-Interscience, 1991, pp. 491-556.

[29] R. Azad, H.R. Shayegh and H. Amiri " A Robust and Efficient Method for Improving Accuracy of License Plate Characters Recognition," International Conference on computer, Information Technology and Digital Media, 2013, pp. 77-82.

[30] Karthik S, H R Mamatha, K Srikanta Murthy, "Kannada Characters Recognition - A Novel Approach Using Image Zoning and Run Length Count", CIIT International journal of digital image processing, volume 3, 2011, pp.1059-1062.

[31] R. Azad, F. Davami and H.R. Shayegh, " Recognition of Handwritten Persian/Arabic Numerals Based on Robust Feature Set and K-NN Classifier," International Journal of Computer & Information Technologies, vol. 1, no. 3, pp. 220-230, 2013.

[32] Vapnik, V., "The Nature of Statistical Learning Theory", New York, Springer-Verlag, 1995.

[33] Joachims, T., "Text categorization with support vector machines:Learning with many relevant features", Technical report, University of Dortmund, 1997, doi: 10.1007/BFb0026683

[34] Wang, T.-Y., Chiang, H.-M., "Fuzzy support vector machine for multi-class text categorization", Information Process and Management, 43, 914–929, 2007, doi: 10.1016/j.ipm.2006.09.011

[35] Salomon, J., "Support vector machines for phoneme classification", M.Sc Thesis, University of Edinburgh, 2001.

[36] R. Azad, B. Azad and I. T. Kazeroni, "Optimized Method for Real-Time Face Recognition System Based on PCA and Multiclass Support Vector Machine", Advances in Computer Science: an International Journal, Vol. 2, Issue 5, No.6, November 2013, pp. 126-132.

[37] Platt, J., Cristianini, N., Shawe-Taylor, J., "Large margin DAGs for multiclass classification", Advances in Neural Information Processing Systems 12. MIT Press, 543–557, 2000.

[38] F. Melgani and L. Bruzzone, "Classification of hyperspectral remote sensing images with support vector machine," IEEE Trans. Geosci. Remote Sens., vol. 42, no. 8, pp. 1778–1790, Aug. 2004, doi: 10.1109/IGARSS.2002.1025088

[39] C.-W.Hsu and C.-J. Lin, "A comparison of methods formulticlass support vector machines," IEEE Trans. Neural Netw., vol. 13, no. 2, pp. 415–425, Mar. 2002.

[40] H. Hamidirad, H. Porgasem, H. Mahdavinasab, "Recognition of Road speed signs for application in intelligent driving," 6th Iranian conference on machine vision and image processing, 2010, pp. 340-345.